\PassOptionsToPackage{table}{xcolor}  
\PassOptionsToPackage{colorlinks=true, linkcolor=blue, citecolor=blue, urlcolor=blue}{hyperref}

\documentclass[letterpaper, 10 pt, conference]{ieeeconf}  

\IEEEoverridecommandlockouts                             

\usepackage{amsmath}
\usepackage{amssymb}

\usepackage[table]{xcolor} 
\usepackage{colortbl}  

\usepackage{graphicx}
\usepackage{svg}
\usepackage[margin=10pt]{subcaption}
\usepackage{wrapfig}
\usepackage{cite}  
\usepackage{url}
\usepackage{array}
\usepackage{booktabs}
\usepackage{multirow}
\usepackage{makecell}
\usepackage{adjustbox}
\usepackage{multicol}

\let\labelindent\relax
\usepackage{enumitem}

\usepackage{algorithm}
\usepackage[noend]{algpseudocode}
\usepackage{xcolor}
\usepackage{tcolorbox}
\usepackage{float}

\usepackage{mathtools}
\usepackage[dvipsnames]{xcolor}
\newtagform{diffnum}[\textcolor{BrickRed}]{(}{)}  
\usepackage{xcolor}
\usepackage{colortbl}

\definecolor{Denim}{HTML}{1F4E79} 
\colorlet{tbmcolor}{Denim}
\newcommand{\tbm}[1]{\textbf{\textcolor{tbmcolor}{#1}}}

\floatstyle{plain}     
\restylefloat{algorithm}

\definecolor{guidelineBlue}{HTML}{2962FF}
\definecolor{guidelineGray}{HTML}{6B6B6B}
\definecolor{guidelineBack}{HTML}{FCFCFC}   
\definecolor{equationGrey}{HTML}{6F6F6F}
\algrenewcommand\algorithmicend{\textcolor{guidelineBlue}{\textbf{end}}}
\algrenewcommand\algorithmicif{\textcolor{guidelineBlue}{\textbf{if}}}
\algrenewcommand\algorithmicthen{\textcolor{guidelineBlue}{\textbf{then}}}
\algrenewcommand\algorithmicelse{\textcolor{guidelineBlue}{\textbf{else}}}
\algrenewcommand\algorithmicfor{\textcolor{guidelineBlue}{\textbf{for}}}
\algrenewcommand\algorithmicdo{\textcolor{guidelineBlue}{\textbf{do}}}
\algrenewcommand\algorithmicwhile{\textcolor{guidelineBlue}{\textbf{while}}}

\algrenewcommand\algorithmiccomment[1]{\hfill\textcolor{guidelineGray}{\footnotesize// #1}}
\tcbset{
  guidelinebox/.style={
    colback=guidelineBack,
    colframe=guidelineBack, 
    boxrule=0pt,
    arc=1.5mm,
    left=1.2em,
    right=1.2em,
    top=0em,
    bottom=0.3em
  }
}
\usepackage{pifont}

\usepackage{url}
\usepackage{hyperref}

\usepackage{url}
\usepackage{hyperref}
\hypersetup{
    colorlinks=true,
    linkcolor=blue,
    citecolor=blue,
    urlcolor=blue
}
\usepackage[
  font      = small,       
  labelfont = small        
]{caption}

\makeatletter
\renewcommand{\@cite}[2]{\textcolor{equationGrey}{[#1]\if@tempswa\,#2\fi}}
\makeatother

\title{\LARGE \bf
GUIDEd Agents: Enhancing Navigation Policies through Task-Specific Uncertainty Abstraction in Localization-Limited Environments
}

\author{Gokul Puthumanaillam$^{1}$, Paulo Padrao$^{2}$, Jose Fuentes$^{3}$, Leonardo Bobadilla$^{3}$, Melkior Ornik$^{1}$
\thanks{$^{1}$University of Illinois Urbana-Champaign. $^{2}$Providence College. $^{3}$Florida International University. Contact emails: {\tt\small gokulp2@illinois.edu, ppadraol@providence.edu, jfuen099@fiu.edu, bobadilla@cs.fiu.edu, mornik@illinois.edu}}%
\thanks{This work was supported in part by NSF grants 2118329, IIS-2024733 and IIS-2331908, the ONR grants N00014-23-1-2789, N00014-25-1-2369, N00014-25-1-2519 and N00014-23-1-2505, the U.S. DoD grant 78170-RT-REP, the Florida Department of Environmental Protection grant INV31, the ARL grant W911NF1920243. This is contribution \#2092 from the Institute of Environment at Florida International University.
}
}

\begin{document}

\maketitle
\thispagestyle{empty}
\pagestyle{empty}

\begin{abstract}
Autonomous robots performing navigation tasks in complex environments often face significant challenges due to uncertainty in state estimation. 
In settings where the robot faces significant resource constraints and accessing high-precision localization comes at a cost, the planner may have to rely primarily on less precise state estimates. 
Our key observation is that different tasks, and different portions of tasks require varying levels of precision in different regions: a robot navigating a crowded space might need precise localization near obstacles but can operate effectively with less precise state estimates in open areas.
In this paper, we present a planning method for integrating task-specific uncertainty requirements directly into navigation policies. We introduce Task-Specific Uncertainty Maps (TSUMs), which abstract the acceptable levels of state estimation uncertainty across different regions. TSUMs align task requirements and environmental features using a shared representation space, generated via a domain-adapted encoder.
Using TSUMs, we propose $\textbf{G}$eneralized $\textbf{U}$ncertainty $\textbf{I}$ntegration for $\textbf{D}$ecision-Making and $\textbf{E}$xecution (GUIDE), a policy-conditioning framework that incorporates these uncertainty requirements into the robot's decision-making process, enabling the robot to reason about the context-dependent value of certainty and adapt its behavior accordingly. 
We show how integrating GUIDE into reinforcement learning frameworks allows the agent to learn navigation policies that effectively balance task completion and uncertainty management without the need for explicit reward engineering. We evaluate GUIDE on a variety of real-world robotic navigation tasks and find that it demonstrates significant improvement in task completion rates compared to baseline methods that do not explicitly consider task-specific uncertainty.
Experiments, code, dataset available at: \texttt{\hyperlink{https://guided-agents.github.io}{https://guided-agents.github.io}}. 

\end{abstract}

\section{Introduction}
In complex environments where robots must balance task completion with resource usage, managing uncertainty in state estimation becomes a critical challenge. 
\begin{figure}[htbp]
    \centering
    \includegraphics[trim=10pt 80pt 50pt 70pt, clip, width=0.35\textwidth]{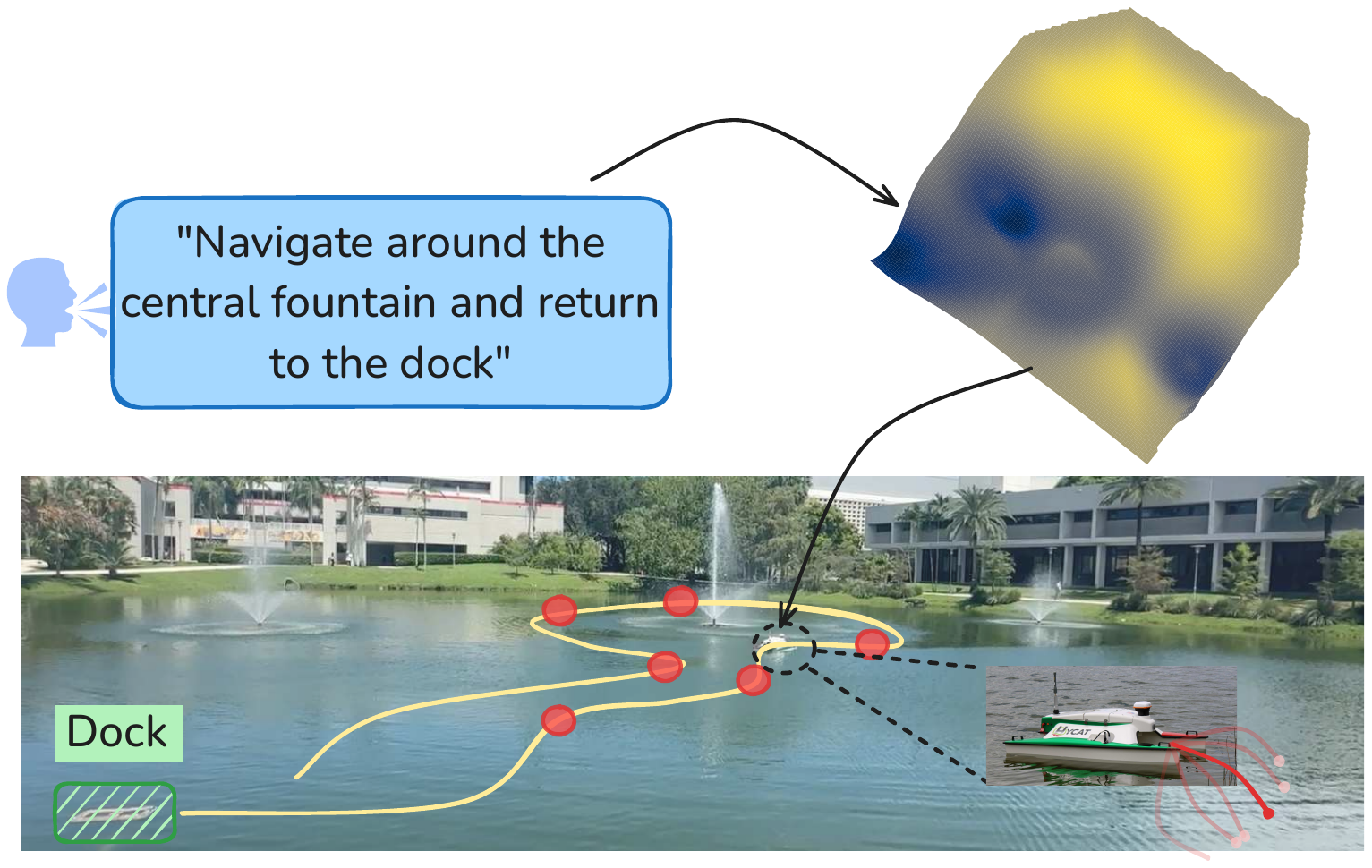}
    \caption{\small Top: The ASV is assigned a navigation task. GUIDE interprets the task and generates a representation highlighting areas where the ASV needs higher positional certainty (dark blue). Bottom: The policy executed by the ASV (white line). The red dots indicate locations where the ASV actively reduces its state estimation uncertainty to satisfy task-specific requirements.}
    \label{fig:teaser}
\end{figure}
Consider an autonomous surface vehicle (ASV) conducting a mapping mission in stealth-critical settings where each GPS fix risks detection, or a ground robot operating in environments where high-precision localization requires costly computation. In such resource-constrained settings, the robot must not only complete its assigned task but also carefully manage when and where to use expensive localization operations. 
Importantly, constantly striving to reduce uncertainty through high-precision localization is neither necessary nor efficient for successful task execution. When operating in regions critical to the task objectives or near obstacles, precise localization becomes crucial for mission success. However, in areas less relevant to the immediate task goals, the robot can often tolerate higher positional uncertainty, allowing it to focus on efficient task execution rather than achieving pinpoint accuracy. This scenario underscores a fundamental observation in robotic navigation with limited access to high-precision localization: the acceptable level of uncertainty at any location is inherently \emph{task-specific} and \emph{context-dependent}.

Traditional approaches to navigation in localization-limited environments often fall into two categories: minimizing uncertainty universally through frequent use of expensive localization \cite{1g, puthumanaillam2024comtraq} or enforcing fixed uncertainty thresholds across the environment \cite{2q, 1w}. While these strategies may be effective in settings where localization resources are abundant, they become impractical when high-precision state estimation comes at a cost. 
Existing methods 
fail to account for how the value of precise localization varies across different phases of the mission. Adapting these methods to localization-limited environments would require extensive reward engineering to capture the complex relationship between task objectives and localization costs.

The key issue lies in the disconnect between task requirements and resource-constrained uncertainty management. Current frameworks treat localization decisions independently from task specifications, leading to either excessive resource consumption or compromised task performance when different regions demand different levels of certainty.

\noindent \tbm{Statement of contributions:} $(i)$ \underline{TSUM}: With the aim of grounding high-level task descriptions and low-level uncertainty handling, we propose the concept of a \emph{\textbf{T}ask-\textbf{S}pecific \textbf{U}ncertainty \textbf{M}ap (TSUM)}. A TSUM serves as an intermediate abstraction that represents the acceptable levels of state estimation uncertainty across different regions of the environment for a specific task. 
We present a CLIP-based approach for computing TSUMs that effectively aligns visual environmental features with textual task specifications in a shared embedding space.
$(ii)$ \underline{GUIDE}: Building upon TSUM, we introduce \emph{\textbf{G}eneralized \textbf{U}ncertainty \textbf{I}ntegration for \textbf{D}ecision-Making and \textbf{E}xecution (GUIDE)}, a policy-conditioning framework that incorporates TSUMs into navigation policies. Our key finding is that by conditioning navigation policies on TSUMs, robots can better reason about the \emph{context-dependent value of certainty}. 
$(iii)$ \underline{GUIDEd agents}: We demonstrate how GUIDE can be integrated into a reinforcement learning framework by adapting the Soft Actor-Critic algorithm, resulting in the \emph{GUIDEd Soft Actor-Critic (G-SAC)}. G-SAC learns navigation policies that effectively balance task completion and uncertainty management without the need for explicit reward engineering. $(iv)$ \underline{Evaluation}: We evaluate GUIDE in real-world in-the-wild deployment using an Autonomous Surface Vehicle in the context of marine autonomy. Our results show significant improvements in task performance when compared to methods that do not explicitly consider task-specific uncertainty requirements. 

\section{Related Works}

\noindent \tbm{Uncertainty Modeling:} Uncertainty management is a fundamental aspect of robot navigation \cite{1a}. Probabilistic techniques \cite{1q,1u} have been widely used to enable localization in stochastic environments. However, these approaches often treat uncertainty uniformly across the environment, without considering how different regions may require varying levels of certainty depending on the task. 
Belief-space planners framed as Partially Observable Markov Decision Processes (POMDPs) further couple estimation with action by optimizing over the distribution, explicitly balancing task progress against the value of information‐gathering \cite{kaelbling1998planning,roy2003coastal}. 
While POMDP approximations improve scalability, they cannot capture location- or context-dependent tolerance. 


\noindent \tbm{RL for Navigation:} Reinforcement Learning (RL) has been successfully applied to robotic navigation tasks, enabling agents to learn navigation policies \cite{1w,1e}. 
While these methods can learn effective policies in controlled environments, they often perform suboptimally in real-world scenarios due to factors like partial observability and dynamic changes \cite{2b, 2g}. 
Recent work has focused on developing navigation policies conditioned on specific goals \cite{3e, 3f} capable of handling objectives, adjusting the robot's behavior according to the task. These approaches often neglect the varying importance of uncertainty management across different tasks and environments.
\begin{figure*}[htbp]
    \centering
    \includegraphics[trim=0pt 0pt 100pt 20pt, clip, width=0.85\textwidth]{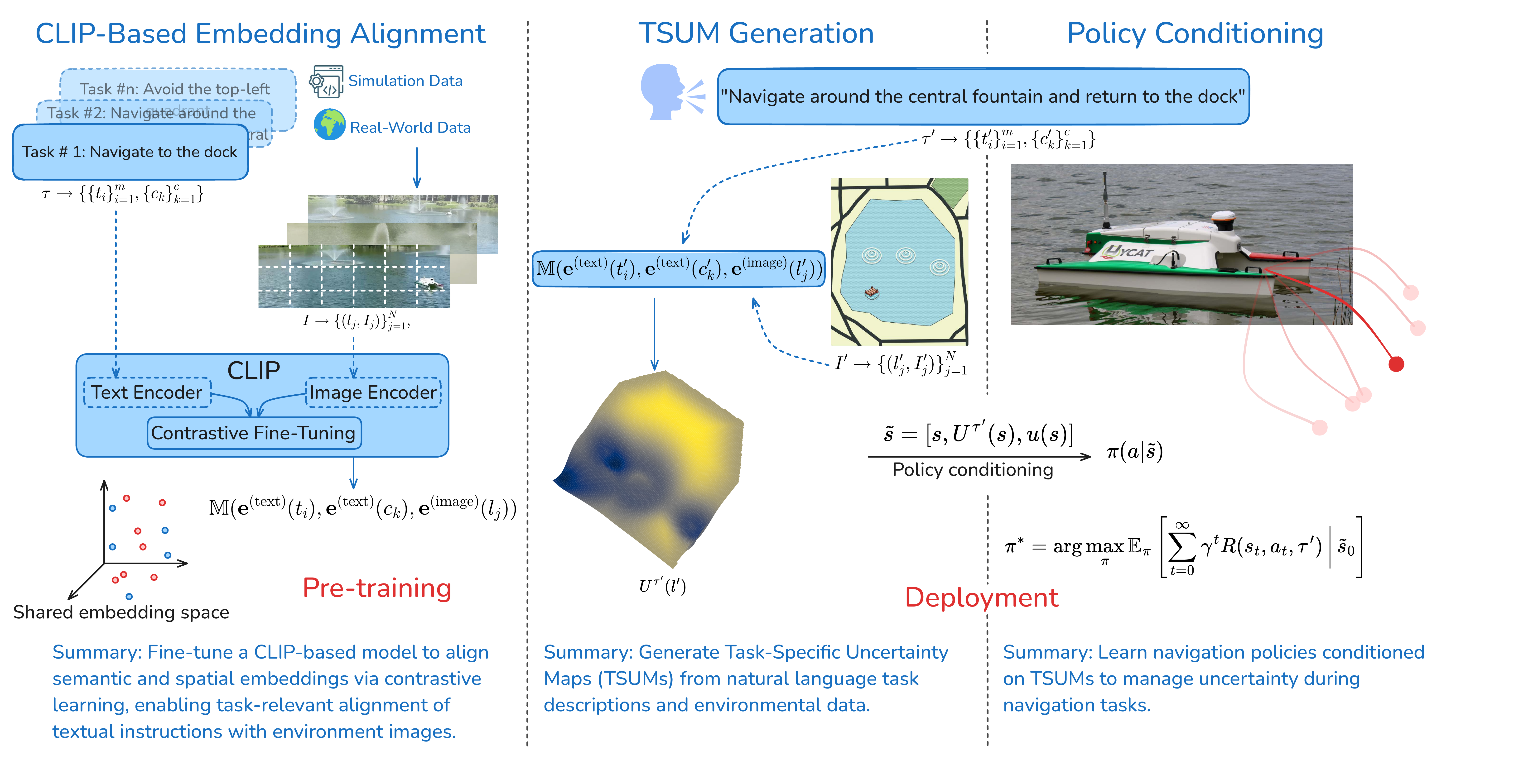}
\caption{\small GUIDE consists of two phases. (i) Pre-training: semantic and spatial embeddings are fine-tuned and aligned using a CLIP-based model. (ii) Deployment: TSUMs are generated from task descriptions and policies are conditioned on the TSUMs to manage uncertainty.}
    \label{fig:architecture}
\end{figure*}

\noindent \tbm{Uncertainty-Aware RL:} Incorporating uncertainty into RL for navigation has been investigated to enhance exploration, robustness, and safety \cite{1e}. Some approaches introduce uncertainty penalization into the reward function \cite{2k, 2l}. Others utilize Bayesian RL methods to model uncertainty in value estimation \cite{2n, 2o}. Bootstrapped ensembles maintain multiple value functions to capture epistemic uncertainty, leading to informed exploration strategies \cite{2t, 2w}. While these methods consider uncertainty, they often do so globally \cite{2q} across the environment and do not tailor to specific navigation tasks or spatial regions. 
This uniform treatment of uncertainty becomes particularly problematic in resource-constrained settings, lacking mechanisms to reason about when consuming these limited resources would provide the most value for task completion.

\noindent \tbm{Risk-Aware Planning:} Risk-aware planning introduces measures to balance performance and safety by considering the risks associated with different actions \cite{3m, 3n}. 
In the context of RL, risk-sensitive approaches adjust the policy to avoid high-risk actions, often through modified reward functions or policy constraints \cite{3s, 3t}. Although effective in managing risk, they apply uniform risk thresholds and do not adapt to the task-specific uncertainty requirements.

\noindent \tbm{Task-Specific Navigation:} Recent works proposed large pre-trained models to interpret complex instructions and generate corresponding navigation behaviors \cite{3u, 3y}. While these approaches enable robots to execute a wider range of tasks, they lack a systematic method for integrating task-specific uncertainty requirements into the navigation policy. This limitation reduces their effectiveness in localization-limited environments where the importance of uncertainty can vary across different regions.

In contrast to prior works, we place our research at the intersection of task-specific, resource-constrained navigation.
GUIDE addresses a specific gap in the literature: \textit{how to complete navigation tasks in environments where high-precision localization is a limited resource that must be strategically utilized}. 
While previous approaches either assume unlimited access to state estimation or treat uncertainty uniformly, GUIDE grounds task requirements, environmental features and uncertainty into the navigation policy. 

\section{The GUIDE Framework}
\label{sec:problem}
\noindent \tbm{Problem setting: }Consider a robot operating in a state space \(\mathcal{S}\), where each state \(s \in \mathcal{S}\) encodes the robot's pose. The robot can execute control inputs from a action space \(\mathcal{A}\). Operating in a localization-limited environment \cite{puthumanaillam2024comtraq}, the robot's default state estimate is subject to uncertainty (e.g., communication constraints or environmental factors). The robot can access high-precision state estimates at significant costs---for instance, a robot conducting stealth-critical missions where each GPS fix risks detection. We assume the availability of a visual representation of the operating environment \emph{a priori}, such as an overhead map. This assumption is justified in real-world applications where geographic information systems provide knowledge of the environment. 
A \emph{navigation task} \(\tau\), specified in natural language, describes the robot's objectives and constraints with respect to environmental features.

\noindent \tbm{Uncertainty representation:} \label{unc}
We represent the robot's pose as a Bayesian belief with covariance $P_t$ maintained by the onboard estimator. To obtain a single, physically meaningful scalar of positional uncertainty, we use the 95\% confidence radius
$u(s_t) \triangleq r_{95}(P_t) \;=\; \sqrt{\chi^2_{2,\,0.95}\,\lambda_{\max}\!\big(P^{(xy)}_t\big)}\,,$
where $\lambda_{\max}$ is the largest eigenvalue of the $(x,y)$ covariance submatrix and $\chi^2_{2,\,0.95}$ is the 95th percentile of the chi-square distribution with two degrees of freedom. The quantity $u(s_t)$ has units of meters and evolves with motion and sensing: dead-reckoning increases $P_t$ through process noise, whereas an exact GPS update contracts $P_t$ via a measurement update. All uncertainty-aware components in our approach are calibrated to this scalar summary.

\noindent \tbm{Objective:} Synthesize a navigation policy \(\pi(a \mid s)\) that enables task completion while explicitly reasoning about the robot's state-estimation uncertainty \(u(s)\). The policy must balance two competing demands: efficient task execution and strategic uncertainty management. 

\noindent \tbm{GUIDE} aims to operationalize this balance by conditioning the navigation policy on a task-specific uncertainty abstraction, enabling the robot to reason \emph{when} and \emph{where} to invest in costly localization. Fig.~\ref{fig:architecture} presents an overview of GUIDE. We now detail the specifics of these components.

\subsection{Task-Specific Uncertainty Map (TSUM)}
To effectively integrate task-specific uncertainty considerations into robotic navigation policies, we introduce the abstraction of \emph{Task-Specific Uncertainty Maps} (TSUMs). A TSUM serves as an intermediate representation that bridges the gap between high-level task specifications and low-level uncertainty management decisions. Formally, a TSUM is a function \( U^\tau: L \rightarrow \mathbb{R}^+ \) that assigns an acceptable level of state estimation uncertainty to each location \( l \in L \) for a given navigation task \( \tau \).
The TSUM \( U^\tau(l) \) is defined as:
\begin{equation} \label{eq:tsum_definition}
U^\tau(l) = w_\Phi\Phi^\tau(l) + w_\mathcal{C}\mathcal{C}^\tau(l) + w_\mathcal{E}\mathcal{E}(l),
\end{equation}
where \(\Phi^\tau(l)\) quantifies the task-specific importance of location \(l\), \(\mathcal{C}^\tau(l)\) represents task constraints at \(l\), and \(\mathcal{E}(l)\) captures environmental factors. The weights \(w_\Phi\), \(w_\mathcal{C}\), and \(w_\mathcal{E}\) control the relative influence of each component, allowing us to balance different aspects of the task and environment. These weights are determined through validation on a held-out dataset of annotated scenarios.

\subsubsection{CLIP-Based TSUM Generation}
To generate TSUMs that effectively capture the relationship between visual features and task requirements, we propose a vision-language approach using the CLIP (Contrastive Language-Image Pre-training) model \cite{w8}. GUIDE processes both visual and textual inputs to produce a spatially-varying uncertainty map. 

\noindent \tbm{Pre-processing:} \textit{(i) Environment representation: }As established in Section \ref{sec:problem}, we assume access to a visual representation of the environment. 
To process this spatial information systematically, we discretize the continuous space \(L\) into a grid of \(N\) image patches \(\{I_1, I_2, \ldots, I_N\}\), where each patch \(I_j\) corresponds to a discrete spatial region \(l_j \subset L\). The discretization parameters are chosen to balance computational efficiency with feature preservation.
\textit{(ii) Task Processing:}
Given a task \(\tau\) specified in natural language, we employ a hierarchical parsing approach to extract structured task representations. We utilize a dependency parser with a context-free grammar \cite{x1} to decompose \(\tau\) into atomic statements that capture objectives and their associated constraints. These atomic statements are organized into \(m\) specifications \(\{t_i\}_{i=1}^m\) and \(c\) auxiliary specifications \(\{c_k\}_{k=1}^c\). 

\noindent \tbm{CLIP Encoding:}
We leverage CLIP's vision-language architecture to project both spatial and semantic information into a shared representational space. The model consists of two parallel encoders:
\begin{equation}
\label{eq:clip_encoders}
    E_\text{text}: \mathcal{T} \rightarrow \mathbb{R}^d, \quad E_\text{image}: \mathcal{I} \rightarrow \mathbb{R}^d
\end{equation}
where \(\mathcal{T}\) and \(\mathcal{I}\) denote the spaces of tokenized text and normalized image patches respectively, and $d$
is chosen to optimize the balance between representational capacity and computational efficiency. These encoders employ self-attention mechanisms and cross-modal transformers to generate context-aware embeddings \(\mathbf{e}^\text{(text)}(t_i), \mathbf{e}^\text{(text)}(c_k) \in \mathbb{R}^d\) for textual components and \(\mathbf{e}^\text{(image)}(l_j) \in \mathbb{R}^d\) for spatial regions.

\noindent \tbm{Cross-Modal Alignment:}
The shared embedding space facilitates direct semantic comparison between specifications and visual features through cosine similarity:
\begin{equation}
    \text{sim}(t, l) = \text{cosine}\big(\mathbf{e}^\text{(text)}(t),\;\mathbf{e}^\text{(image)}(l)\big)
\end{equation}
This metric quantifies the semantic relevance between spatial regions and task components, enabling the construction of spatially-varying relevance maps. While pre-trained CLIP provides general-purpose understanding, domain-specific applications would require additional fine-tuning
(Sec. \ref{sec:implementation}).

\begin{figure*}[htbp]
    \centering
    \includegraphics[trim=130pt 120pt 60pt 20pt, clip, width=0.85\textwidth]{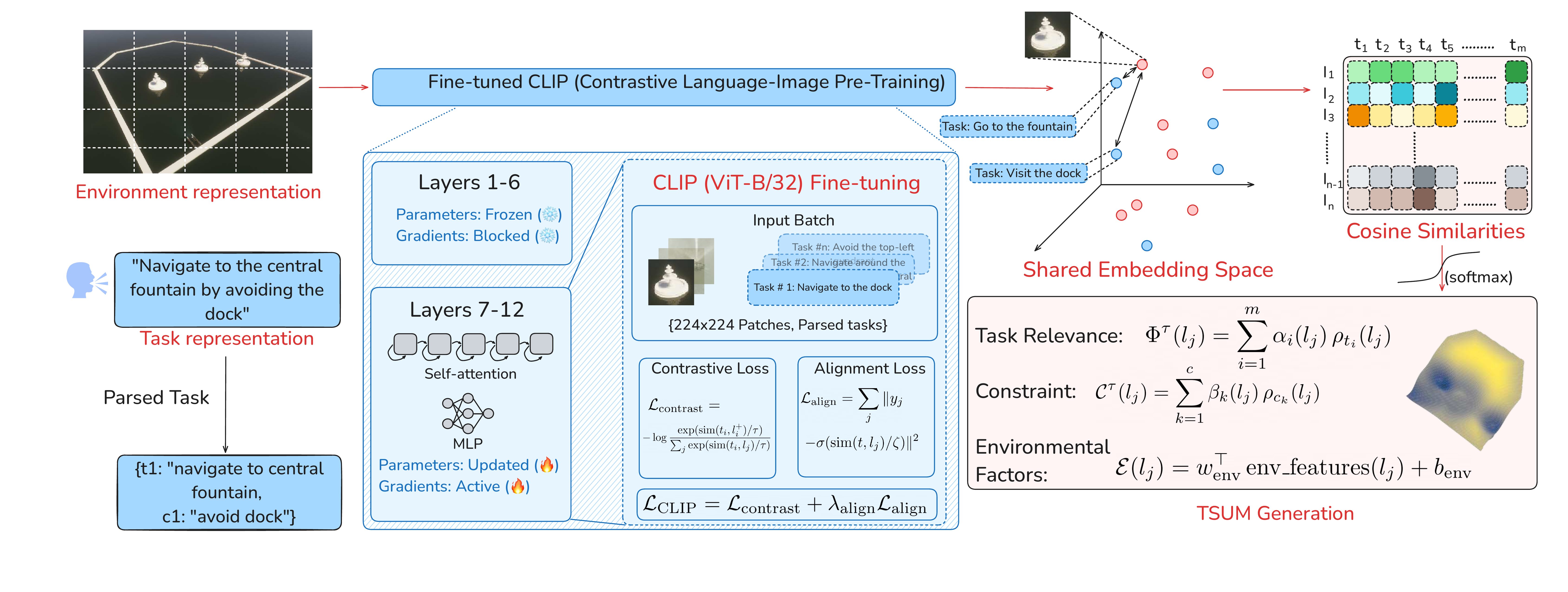}
\caption{\small CLIP-based TSUM generation pipeline. Environment imagery and task description are encoded by a fine-tuned CLIP; their joint embeddings produce task-relevance and constraint maps that, fused with environment priors, yield the final TSUM.}
    \label{fig:clip_pipeline}
\end{figure*}

\subsubsection{Computing Task Relevance and Constraints}
Having established our CLIP-based pipeline, we now detail how we compute each component of the TSUM. 
Using the encoder functions defined in Equation~(\ref{eq:clip_encoders}), we obtain embeddings (\(\mathbf{e}^\text{(text)}(t_i), \mathbf{e}^\text{(text)}(c_k) \in \mathbb{R}^d\) for text and \(\mathbf{e}^\text{(image)}(l_j) \in \mathbb{R}^d\) for image patches) that capture the semantic features of both textual descriptions and visual characteristics in a shared \(d\)-dimensional space. This shared representation allows us to meaningfully compare how well visual features at each location align with the requirements expressed in the task.

\noindent \tbm{Task Relevance Function \(\Phi^\tau(l)\):}
To compute the task relevance at each location, we first calculate similarity scores between each subtask and location:
$\rho_{t_i}(l_j) = \text{cosine}\big(\mathbf{e}^\text{(text)}(t_i),\;\mathbf{e}^\text{(image)}(l_j)\big).$
These raw similarities are converted to attention weights
$\alpha_i(l_j) = {\exp\big(\rho_{t_i}(l_j)\big)}/{\sum_{i'=1}^m \exp\big(\rho_{t_{i'}}(l_j)\big)}.$ 
The task relevance function \(\Phi^\tau(l_j)\) is then computed as a weighted sum
\begin{equation}
\Phi^\tau(l_j) = \sum_{i=1}^m \alpha_i(l_j)\,\rho_{t_i}(l_j). 
\end{equation}
This formulation creates a natural prioritization mechanism--locations that strongly match with important subtasks receive higher relevance scores. The attention weights \(\alpha_i\) ensure that the most pertinent subtasks for each location dominate the relevance calculation, while the weighted sum combines evidence from all subtasks proportional to their importance.

\noindent \tbm{Constraint Function \(\mathcal{C}^\tau(l)\):}
For each constraint \(c_k\),
$\rho_{c_k}(l_j) = \text{cosine}\big(\mathbf{e}^\text{(text)}(c_k),\;\mathbf{e}^\text{(image)}(l_j)\big).$
Constraint attention weights are computed as
$\beta_k(l_j) = {\exp\big(\rho_{c_k}(l_j)\big)}/{\sum_{k'=1}^c \exp\big(\rho_{c_{k'}}(l_j)\big)}. $
The final constraint function aggregates these similarities
\begin{equation}
\mathcal{C}^\tau(l_j) = \sum_{k=1}^c \beta_k(l_j)\,\rho_{c_k}(l_j).
\end{equation}
Higher values of \(\mathcal{C}^\tau(l_j)\) indicate locations where constraints are relevant and uncertainty management is critical.

\noindent \tbm{Environmental Factors \(\mathcal{E}(l)\):}
While CLIP embeddings capture visual features, certain navigation-relevant attributes may not be directly observable from imagery. We incorporate these factors through a vector of environmental features \(\text{env\_features}(l_j)\), which encodes spatial properties. These features are mapped to a scalar value through
\begin{equation}
\mathcal{E}(l_j) = w_\text{env}^\top\,\text{env\_features}(l_j) + b_\text{env},
\end{equation}
where \(w_\text{env}\) and \(b_\text{env}\) are optimized during training.

The final TSUM value at each location is computed by combining these components according to Equation~(\ref{eq:tsum_definition}). The complete pipeline is illustrated in Fig.~\ref{fig:clip_pipeline}.

\subsubsection{Implementation Recipe} \label{sec:implementation}
We now present the practical details of our domain-specific implementation.
\noindent \tbm{Dataset}
Our dataset consists of 2500 annotated overhead images (2000 images for training and 500 for validation) collected from real-world environments and high-fidelity simulations, covering a diverse range of navigation scenarios. Each image is discretized to 224 \(\times\) 224 pixels at 0.5 meters per pixel resolution, chosen to align with CLIP's architecture.
Each patch is annotated with
$\mathcal{D} = \{(I_j, l_j, y_j^t, y_j^c, f_j)\}_{j=1}^N$,
where \(I_j\) is the image patch, \(l_j\) represents location coordinates, \(y_j^t\) and \(y_j^c\) are binary relevance labels for a predefined set of subtasks and constraints, and \(f_j\) contains environmental features (e.g., bathymetry). 

\noindent \tbm{CLIP Fine-tuning:} 
We adopt a selective fine-tuning approach to adapt the pre-trained CLIP (ViT-B/32) model while preserving its general vision-language understanding. Specifically, we freeze the first six transformer layers to maintain CLIP's foundational cross-modal understanding, while fine-tuning the remaining six layers to adapt to navigation-specific concepts. We optimize the loss function:
\begin{equation}
\mathcal{L}_\text{CLIP} = \mathcal{L}_\text{cont.} + \lambda_\text{align}\mathcal{L}_\text{align},
\end{equation}
\(\mathcal{L}_\text{cont.}=-\log({\exp(\text{sim}(t_i, l_i^+)/\tau)}/{\sum_{j}\exp(\text{sim}(t_i, l_j)/\tau)})\) is the standard CLIP contrastive loss,
and $\mathcal{L}_\text{align} = \sum_j\|y_j - \sigma(\text{sim}(t, l_j)/\zeta)\|^2$ is an alignment loss that encourages consistency between expert labels and embedding similarities. 
\(\zeta\) is a temperature parameter set to 0.07, and \(\sigma\) is the sigmoid function. We freeze the first six transformer layers of CLIP and fine-tune the remaining layers for 10 epochs using AdamW optimizer with a learning rate of 1e-5.

\noindent \tbm{Training Procedure:}
The training process involves:
\begin{enumerate}[label=(\roman*), nosep]
\item Fine-tune CLIP using \(\mathcal{L}_\text{CLIP}\) as described above.
\item Train \(w_\text{env}\) and \(b_\text{env}\) by minimizing
$\mathcal{L}_\text{env} = \sum_j\|\mathcal{E}(l_j) - \hat{e}_j\|^2,$
where \(\hat{e}_j\) represents ground truth annotations.
\item Optimize the weights \(w_\Phi\), \(w_{\mathcal{C}}\), and \(w_{\mathcal{E}}\) using a validation set of TSUMs
$\mathcal{L}_\text{TSUM} = \sum_j\|U^\tau(l_j) - U^\tau_\text{ref}(l_j)\|^2.$
\end{enumerate}

The reference TSUMs (\(U^\tau_\text{ref}\)) were generated using a combination of automated and manual annotation processes. We utilized a Unity-based simulation environment to generate initial uncertainty requirements based on task execution traces. 
We employ early stopping based on validation performance with a patience of 5 epochs. Data augmentation includes random rotations and flips of the image patches. The pipeline is implemented in PyTorch \texttt{TorchScript} for optimal GPU utilization. 
Training is performed on distributed NVIDIA A100 GPUs using PyTorch's \texttt{DistributedDataParallel}, with simulation and data generation parallelized across multiple nodes.

\subsection{Conditioning Navigation Policies Using TSUMs}
To enable robots to manage uncertainty in a task-aware manner, we propose conditioning the navigation policy on the TSUM. 
We augment the state representation (\( \tilde{s} \)) to include both the acceptable uncertainty levels from the TSUM and the robot's current estimation of its state uncertainty:
\begin{equation}
\tilde{s} = [s, U^\tau(s), u(s)],
\end{equation}
where \( s \) is the original state representing the robot's observation of the environment, \( U^\tau(s) \) is derived from the TSUM, and \( u(s) \) is the robot's current state estimation uncertainty.

By conditioning the policy \( \pi(a|\tilde{s}) \) on the augmented state \( \tilde{s} \), the agent makes decisions based not only on the environmental state but also on the acceptable, actual and future uncertainty levels at each location. This allows the agent to adjust its actions to meet the task-specific uncertainty requirements. The RL problem is formulated in the augmented state space. The objective is to learn an optimal policy \( \pi^*(a|\tilde{s}) \) that maximizes the expected cumulative reward:
\begin{equation}
\pi^* = \arg\max_\pi \mathbb{E}_{\pi}\left[ \sum_{t=0}^\infty \gamma^t R(s_t, a_t, \tau) \,\Big|\, \tilde{s}_0 \right],
\end{equation}
where \( \gamma \in [0,1) \) is the discount factor, and \(R_\tau =R(s_t, a_t, \tau) \) is the task-specific reward function.
Standard RL algorithms can be employed to solve this optimization problem in the augmented state space. In the next section, we show how a standard RL algorithm can be adapted using our framework. 

\subsubsection{GUIDEd SAC}
While any RL algorithm can be used to learn the optimal policy,
we adopt the Soft Actor-Critic (SAC) algorithm \cite{3l} due to its sample efficiency and robustness. Our adapted version, referred to as \emph{GUIDEd SAC}, conditions the policy and value function on the augmented state \( \tilde{s} \). 
In SAC framework, the objective is to maximize the expected cumulative reward augmented by an entropy term:
\begin{equation}
J(\pi) = \mathbb{E}_{\pi}\left[ \sum_{t=0}^\infty \gamma^t \left( R_\tau + \alpha \mathcal{H}(\pi(\cdot|\tilde{s}_t)) \right) \,\Big|\, \tilde{s}_0 \right],
\end{equation}
\noindent where \( \mathcal{H}(\pi(\cdot|\tilde{s}_t)) = -\mathbb{E}_{a_t \sim \pi(\cdot|\tilde{s}_t)}\left[ \log \pi(a_t|\tilde{s}_t) \right] \) is the entropy at state \( \tilde{s}_t \), and \( \alpha \) is the temperature parameter balancing exploration and exploitation.
GUIDEd SAC maintains parameterized function approximators for the policy \( \pi_\theta(a|\tilde{s}) \) and the soft Q-value functions \( Q_{\phi_1}(\tilde{s}, a) \) and \( Q_{\phi_2}(\tilde{s}, a) \), where \( \theta \) and \( \phi_i \) denote the parameters of the policy and value networks, respectively.
The soft Q-value networks \( Q_{\phi_i}(\tilde{s}, a) \) are updated by minimizing the soft Bellman residual:
\begin{equation} \label{eq:q_update}
\mathcal{L}_Q(\phi_i) = \mathbb{E}_{(\tilde{s}_t, a_t, r_t, \tilde{s}_{t+1}) \sim \mathcal{D}} \left[ \left( Q_{\phi_i}(\tilde{s}_t, a_t) - y_t \right)^2 \right],
\end{equation}
where \( \mathcal{D} \) is the replay buffer, and target \( y_t \) is computed as:
\begin{equation} \label{eq:q_target}
y_t = r_t + \gamma \left( \min_{i=1,2} Q_{\bar{\phi}_i}(\tilde{s}_{t+1}, a_{t+1}) - \alpha \log \pi_\theta(a_{t+1}|\tilde{s}_{t+1}) \right),
\end{equation}
with \( a_{t+1} \sim \pi_\theta(\cdot|\tilde{s}_{t+1}) \) and \( Q_{\bar{\phi}_i} \) being the target Q-value networks with delayed parameters for stability.
The policy network \( \pi_\theta(a|\tilde{s}) \) is updated by minimizing:
\begin{equation} \label{eq:policy_update}
\mathcal{L}_\pi(\theta) = \mathbb{E}_{\tilde{s}_t \sim \mathcal{D}} \left[ \mathbb{E}_{a_t \sim \pi_\theta(\cdot|\tilde{s}_t)} \left[ \alpha \log \pi_\theta(a_t|\tilde{s}_t) - Q_{\phi}(\tilde{s}_t, a_t) \right] \right],
\end{equation}
where \( Q_{\phi}(\tilde{s}_t, a_t) = \min_{i=1,2} Q_{\phi_i}(\tilde{s}_t, a_t) \).
The temperature parameter \( \alpha \) is adjusted by minimizing:
\begin{equation} \label{eq:alpha_update}
\mathcal{L}(\alpha) = \mathbb{E}_{a_t \sim \pi_\theta(\cdot|\tilde{s}_t)} \left[ -\alpha \left( \log \pi_\theta(a_t|\tilde{s}_t) + \bar{\mathcal{H}} \right) \right],
\end{equation}
where \( \bar{\mathcal{H}} \) is the target entropy.
Algorithm~\ref{alg:guided_sac} summarizes the GUIDEd SAC algorithm.


\begin{algorithm}[H]
\caption{\small GUIDEd SAC Algorithm}
\label{alg:guided_sac}
\begin{algorithmic}[1]
\small
\State \textcolor{guidelineBlue}{\textbf{Initialize}} policy network \( \pi_\theta(a|\tilde{s}) \), Q-value networks \( Q_{\phi_1} \), \( Q_{\phi_2} \), target Q-value networks \( Q_{\bar{\phi}_1} \), \( Q_{\bar{\phi}_2} \), temperature parameter \( \alpha \), and replay buffer \( \mathcal{D} \).
\For{each environment interaction step}
    \State Obtain \( s_t \), \( U^\tau(s_t) \), \( u(s_t) \).
    \State Form augmented state \( \tilde{s}_t = [s_t, U^\tau(s_t), u(s_t)] \).
    \State Sample action \( a_t \sim \pi_\theta(\cdot|\tilde{s}_t) \).
    \State Execute action \( a_t \), observe \( r_t \) and \( s_{t+1} \).
    \State Compute \( \tilde{s}_{t+1} = [s_{t+1}, U^\tau(s_{t+1}), u(s_{t+1})] \).
    \State Store transition \( (\tilde{s}_t, a_t, r_t, \tilde{s}_{t+1}) \) in replay buffer.
\EndFor
\For{each gradient step}
    \State Sample minibatch of transitions from \( \mathcal{D} \).
    \State Update \( Q_{\phi_i} \) by minimizing \( \mathcal{L}_Q(\phi_i) \) (\textcolor{equationGrey}{Eq.~\eqref{eq:q_update}}).
    \State Update \( \pi_\theta \) by minimizing \( \mathcal{L}_\pi(\theta) \) (\textcolor{equationGrey}{Eq.~\eqref{eq:policy_update}}).
    \State Adjust \( \alpha \) by minimizing \( \mathcal{L}(\alpha) \) (\textcolor{equationGrey}{Eq.~\eqref{eq:alpha_update}}).
    \State Update target Q networks: \( \bar{\phi}_i \leftarrow \tau \phi_i + (1 - \tau) \bar{\phi}_i \).
\EndFor
\end{algorithmic}
\end{algorithm}


\begin{table*}[t!]
\captionsetup{skip=2pt} 
\centering
\begingroup
\setlength{\aboverulesep}{0.15ex}
\setlength{\belowrulesep}{0.15ex}
\renewcommand{\arraystretch}{0.92} 
\begin{adjustbox}{max width=\textwidth}
\arrayrulecolor{black!30}
\scriptsize
\begin{tabular}{p{0.30\textwidth} l@{\hskip 5pt}|@{\hskip 5pt}c c c c c c | c >{\columncolor{white}}c}
\toprule
\textbf{Tasks} & \textbf{Metric} & \textbf{SAC} & \textbf{SAC-P} & \textbf{B-SAC} & \textbf{CVaR} & \textbf{RAA} & \textbf{HEU} & \textbf{G-PPO} & \cellcolor{cyan!10} \textbf{G-SAC} \\
\midrule
\textbf{Goal reaching: waypoint} & TCR (\%)~($\uparrow$) & 67.2\% & 82.1\% & 75.9\% & 68.8\% & 35.3\% & 71.3\% & 83.8\% & \cellcolor{cyan!10}95.7\% \\ \textcolor{guidelineBlue}{\texttt{visit [coordinate]}} & Reward (avg)~($\uparrow$) & 186.2 & 260.4 & 249.5 & 184.8 & 26.9 & 176.1 & 319.0 & \cellcolor{cyan!10}429.2 \\
\midrule
\textbf{Goal reaching: context} & TCR (\%)~($\uparrow$) & 68.9\% & 84.3\% & 74.2\% & 66.8\% & 32.7\% & 73.5\% & 81.7\% & \cellcolor{cyan!10}91.5\% \\
\textcolor{guidelineBlue}{\texttt{navigate to dock}} & Reward (avg)~($\uparrow$) & 144.1 & 241.8 & 189.2 & 134.6 & 53.5 & 137.5 & 308.3 & \cellcolor{cyan!10}400.2 \\
\midrule
\textbf{Avoid tasks} & TCR (\%)~($\uparrow$) & 71.3\% & 83.2\% & 79.6\% & 78.4\% & 51.3\% & 62.4\% & \cellcolor{cyan!10}89.6\% & 87.4\% \\
\textcolor{guidelineBlue}{\texttt{avoid the central fountain}} & Reward (avg)~($\uparrow$) & 177.8 & 199.2 & 277.6 & 220.4 & 107.8 & 174.4 & 437.6 & \cellcolor{cyan!10}480.2 \\
\midrule
\textbf{Perimeter tasks} & TCR (\%)~($\uparrow$) & 44.3\% & 51.6\% & 56.3\% & 41.6\% & 39.8\% & 49.6\% & 74.9\% & \cellcolor{cyan!10}85.6\% \\
\textcolor{guidelineBlue}{\texttt{go around the left fountain}} & Reward (avg)~($\uparrow$) & 84.4 & 132.8 & 170.4 & 32.8 & 111.2 & 146.8 & 449.2 & \cellcolor{cyan!10}599.6 \\
\midrule
\textbf{Explore tasks} & TCR (\%)~($\uparrow$) & 88.6\% & 92.4\% & 87.6\% & 84.9\% & 71.3\% & 88.7\% & 93.8\% & \cellcolor{cyan!10}97.3\% \\
\textcolor{guidelineBlue}{\texttt{explore top-right quadrant}} & Reward (avg)~($\uparrow$) & 486 & 474 & 526 & 449 & 402 & 537 & 548 & \cellcolor{cyan!10}595 \\
\midrule
\textbf{Restricted areas} & TCR (\%)~($\uparrow$) & 70.8\% & 82.1\% & 78.4\% & 71.2\% & 51.1\% & 63.9\% & \cellcolor{cyan!10}90.0\% & 88.6\% \\
\textcolor{guidelineBlue}{\texttt{visit dock, avoid top-right quadrant}} & Reward (avg)~($\uparrow$) & 266.4 & 306.8 & 307.2 & 269.6 & 201.4 & 261.2 & 570.0 & \cellcolor{cyan!10}634.6 \\
\midrule
\textbf{Multi-goal tasks} & TCR (\%)~($\uparrow$) & 31.3\% & 42.9\% & 37.7\% & 30.9\% & 19.5\% & 42.1\% & 72.8\% & \cellcolor{cyan!10}81.7\% \\
\textcolor{guidelineBlue}{\texttt{combination of tasks (see Fig.~3)}} & Reward (avg)~($\uparrow$) & 124.4 & 135.2 & 122.4 & 129.2 & 100.4 & 155.2 & 423.6 & \cellcolor{cyan!10} 510.4 \\
\bottomrule
\end{tabular}
\end{adjustbox}
\endgroup
\caption{\small Task-completion rate (TCR) and mean episode reward for GUIDE vs baselines, averaged over 50 trials per task type. Task completion criteria: (i) goal-reaching tasks: within a 1.5m radius of the target, (ii) perimeter/exploration tasks: remain within a 3.5 m corridor of the path, (iii) avoid tasks: keep a 3 m distance. TCR is the fraction of these criteria met.}
\label{tab:comparison}
\end{table*}
\section{Experiments}
\textit{Motivating scenario: }Consider a stealth mission where an autonomous surface vehicle (ASV) must navigate through previously unexplored waters. The ASV can either rely on noisy sensors or activate high-precision GPS. Each GPS fix incurs a penalty because it increases the chance of detection, yet completing the mission requires maintaining sufficient localization accuracy in critical regions.

\begin{figure*}[ht!]
    \centering
    \begin{minipage}[b]{0.33\textwidth}  
        \centering
        \includegraphics[width=0.98\textwidth, trim={330pt 260pt 120pt 100pt}, clip]{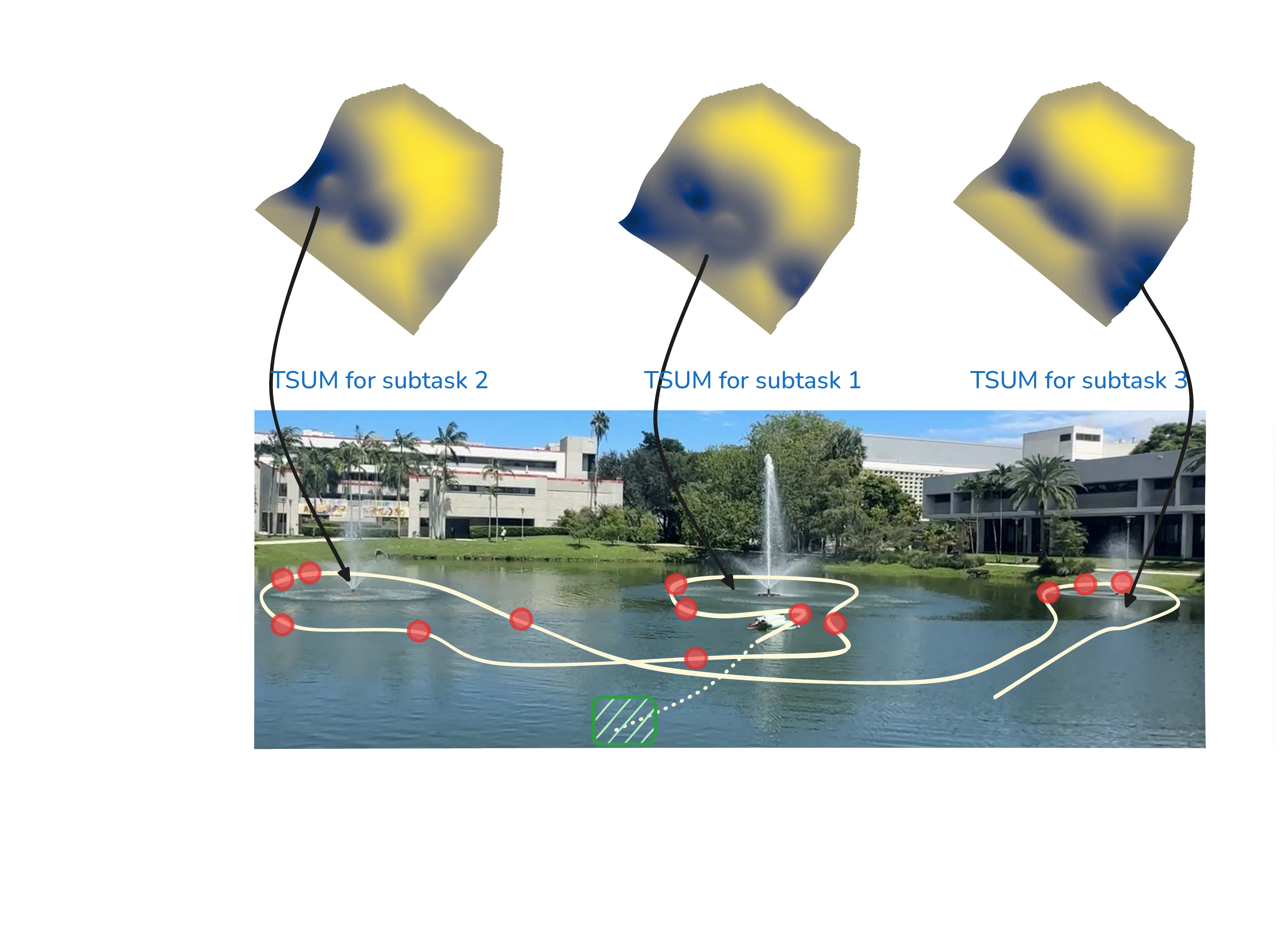}
        \caption*{\small GUIDEd SAC}
    \end{minipage}%
    \begin{minipage}[b]{0.33\textwidth}  
        \centering
        \includegraphics[width=0.99\textwidth, trim={330pt 285pt 120pt 100pt}, clip]{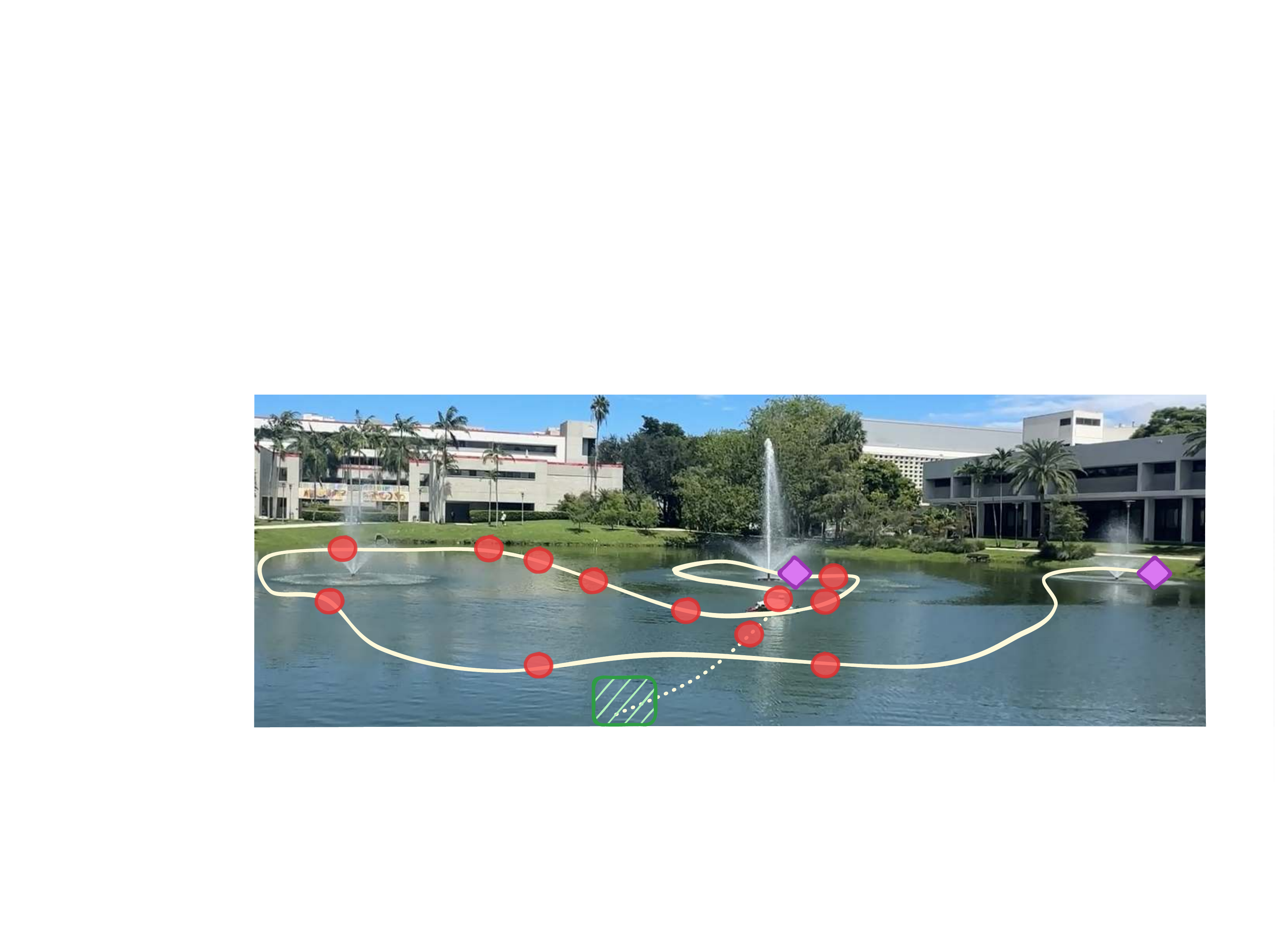}
        \caption*{\small SAC}
    \end{minipage}%
    \begin{minipage}[b]{0.33\textwidth}  
        \centering
                \includegraphics[width=0.99\textwidth, trim={330pt 265pt 120pt 100pt}, clip]{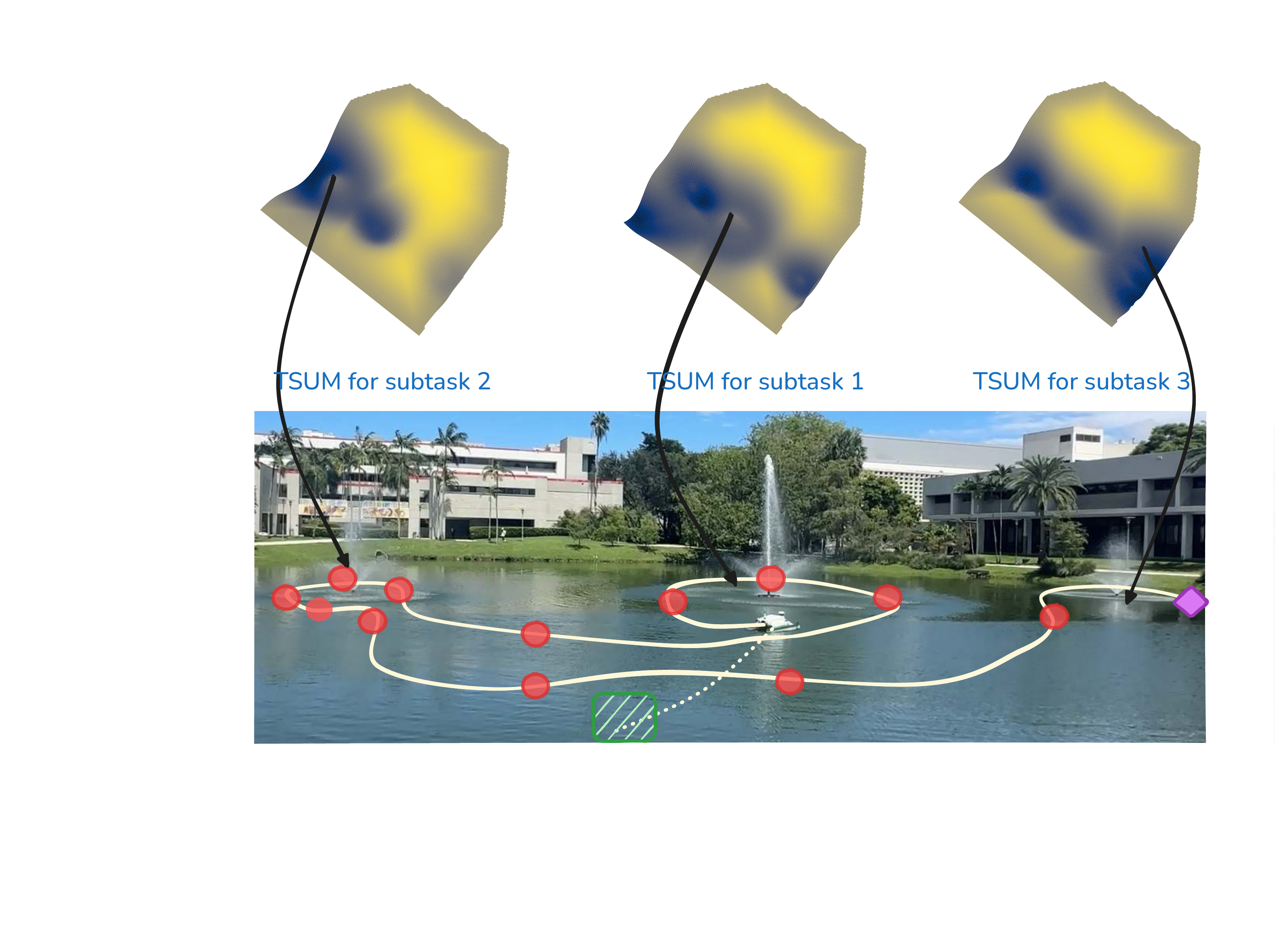}
        \caption*{\small GUIDEd PPO}
    \end{minipage}
    \caption{\small Comparison of navigation trajectories for the task:  \textcolor{guidelineBlue}{\texttt{Start at the dock, navigate around the central fountain, then around the left fountain, and finally around the right fountain.}} Red dots mark locations where it reduced uncertainty, pink diamonds indicate collision points and the green rectangle marks the dock.}
    \label{fig:results1}
\end{figure*}

\subsection{Experimental Setup}

\noindent \tbm{Environment:}
The ASV operates in an open lake characterized by environmental variability and human-induced disturbances. The environment features fountains that serve as obstacles and introduce unmodeled water disturbances that create both physical challenges for navigation and contribute to the overall uncertainty in state estimation.

\noindent \tbm{Hardware platform:}
Experiments use a SeaRobotics Surveyor ASV equipped with GPS, IMU and a YSI EXO2 multiparameter water-quality sonde. The robot exposes (i) the pose $(\hat{x}_t,\hat{y}_t,\hat{\psi}_t)$ and velocities $(\hat{v}_t,\hat{\omega}_t)$, (ii) a switchable localization mode, and (iii) setpoint interfaces for propulsion and steering tracked by the onboard autopilot. 
The action space of the ASV is \( a_t = (\lambda_t, \alpha_t, \eta_t) \), where \( \lambda_t \in [0, 85]\)Nm is the propulsion (throttle), \( \alpha_t \in [0, 2\pi]\) the steering/rudder angle, and \( \eta_t \) the discrete localization mode selector. 

\noindent \tbm{Policy inputs/outputs at each decision step:}
At each decision step $t$, the policy consumes a compact observation vector
$s_t^{\text{num}}=\big[\hat{x}_t,\hat{y}_t,\hat{\psi}_t,\hat{v}_t,\hat{\omega}_t,\; d_{\text{goal}},\theta_{\text{goal}},\; u(s_t),\; U^\tau(s_t)\big],$ where $d_{\text{goal}}$ and $\theta_{\text{goal}}$ denote the Euclidean distance and bearing to the goal.
augmented by a small local TSUM map crop centered at $(\hat{x}_t,\hat{y}_t)$ (encoded and concatenated with $s_t^{\text{num}}$). The three policy outputs are
(i) $\lambda_t$ (throttle setpoint),
(ii) $\alpha_t$ (rudder setpoint), and
(iii) $\eta_t\in\{\text{noisy},\text{exact}\}$.

\noindent \tbm{Uncertainty tracking (runtime):}
We summarize positional belief uncertainty as detailed in \ref{unc}. This is computed from the estimator covariance \(P_t\). The policy thus operates on the augmented state $\tilde{s}$ and learns when to invoke exact localization versus proceeding under noisier estimates.

\noindent \tbm{Position Estimation Modes:}
To represent the operational challenges of managing uncertainty during navigation, we model two modes of position estimation \cite{puthumanaillam2024comtraq}: (i) \textit{Noisy Localization}: By default, the ASV estimates its position using the IMU and EXO2 sensors, which results in a less accurate position estimate. This mode represents the low-cost but high-uncertainty estimate. 
(ii) \textit{Exact Localization}: The ASV can request exact position data from GPS thereby incurring a higher resource cost.


\subsection{Baselines and Ablations}
\noindent We compare GUIDE against baselines that handle uncertainty differently. They interpret tasks using the same parser, converting them into algorithm-specific representations:
\begin{enumerate}[label=(\roman*), nosep]
\item \textit{Standard RL without TSUMs}: We train SAC {\cite{3l}} using the original state \( s \) to assess the importance of TSUMs.
\item \textit{GUIDEd PPO (G-PPO)}: We implement GUIDEd PPO \cite{3h}, incorporating TSUMs and uncertainty, to examine the effect of the underlying RL algorithm.
\item \textit{RL with Uncertainty Penalization (SAC-P)}: We modify the reward function in SAC to penalize high uncertainty \cite{reb1}: \( R_{\text{SAC-P}} = R_{\text{base}} - \zeta u(s) \), where \( R_{\text{base}} \) includes task and localization costs and \( \zeta \) is a weighting factor. This tests traditional reward shaping versus GUIDE.

\item \textit{Bootstrapped Uncertainty-Aware RL (B-SAC)}: We implement Bootstrapped SAC \cite{3i} to estimate epistemic uncertainty which guides exploration.

\item \textit{Heuristic Policy (HEU)}: A hand-crafted policy where the agent plans using SAC but switches to exact position estimation near obstacles or task-critical regions.


\item \textit{Risk-Aware RL (CVaR)}: We use a CVaR-based RL algorithm \cite{3j}, optimizing a risk-sensitive objective to focus on worst-case scenarios.

\item \textit{Uncertainty-Aware Planning (RAA)}: We employ Risk-Aware A* \cite{3k} that minimize collision probabilities, aiming for overall uncertainty minimization.
\end{enumerate}
All baselines, except HEU, use standard implementations lightly adapted to our observation/action spaces.

\subsection{Results and Discussions}

\noindent \tbm{Impact of TSUMs:}
To assess the significance of TSUMs, we compare the standard SAC to G-SAC. As shown in Table~\ref{tab:comparison}, conditioning on TSUMs significantly enhances performance across all tasks. Without TSUMs, SAC cannot manage positional uncertainty in a task-specific manner, leading to suboptimal decisions and lower rewards.
Fig.~\ref{fig:results1} illustrates this difference qualitatively. 
We observe that SAC often overuses high-precision localization in areas where it is unnecessary, incurring additional costs without significant benefits. 
In critical regions requiring precise navigation, SAC fails to reduce uncertainty, leading to collisions. 
In contrast, the G-SAC uses TSUMs to adapt its uncertainty management, switching to GPS in areas where the TSUM indicates low acceptable uncertainty, enabling the ASV to navigate safely around obstacles and complete tasks efficiently.

\noindent \tbm{Effect of RL Algorithm:}
We investigate the impact of the underlying RL algorithm by comparing G-PPO to G-SAC.
This difference in the performance can be attributed to factors inherent to the algorithms. G-SAC, based on the SAC framework, is an off-policy method that leverages entropy regularization to encourage exploration while maintaining stability. Its off-policy nature allows for efficient sample utilization, which is particularly beneficial in continuous action spaces and when data collection is costly or limited.
In contrast, PPO relies on proximal updates to prevent large policy shifts. While PPO is stable, it can be less sample-efficient, as it requires new data for each policy update and may not explore as effectively in complex environments.
Our empirical results suggest that the off-policy efficiency and exploration capabilities of G-SAC make it better suited. 
\begin{figure}[ht!]
    \centering
    \begin{minipage}{\columnwidth}
        \centering
        \includegraphics[trim=45pt 130pt 20pt 30pt, clip, width=0.6\columnwidth]{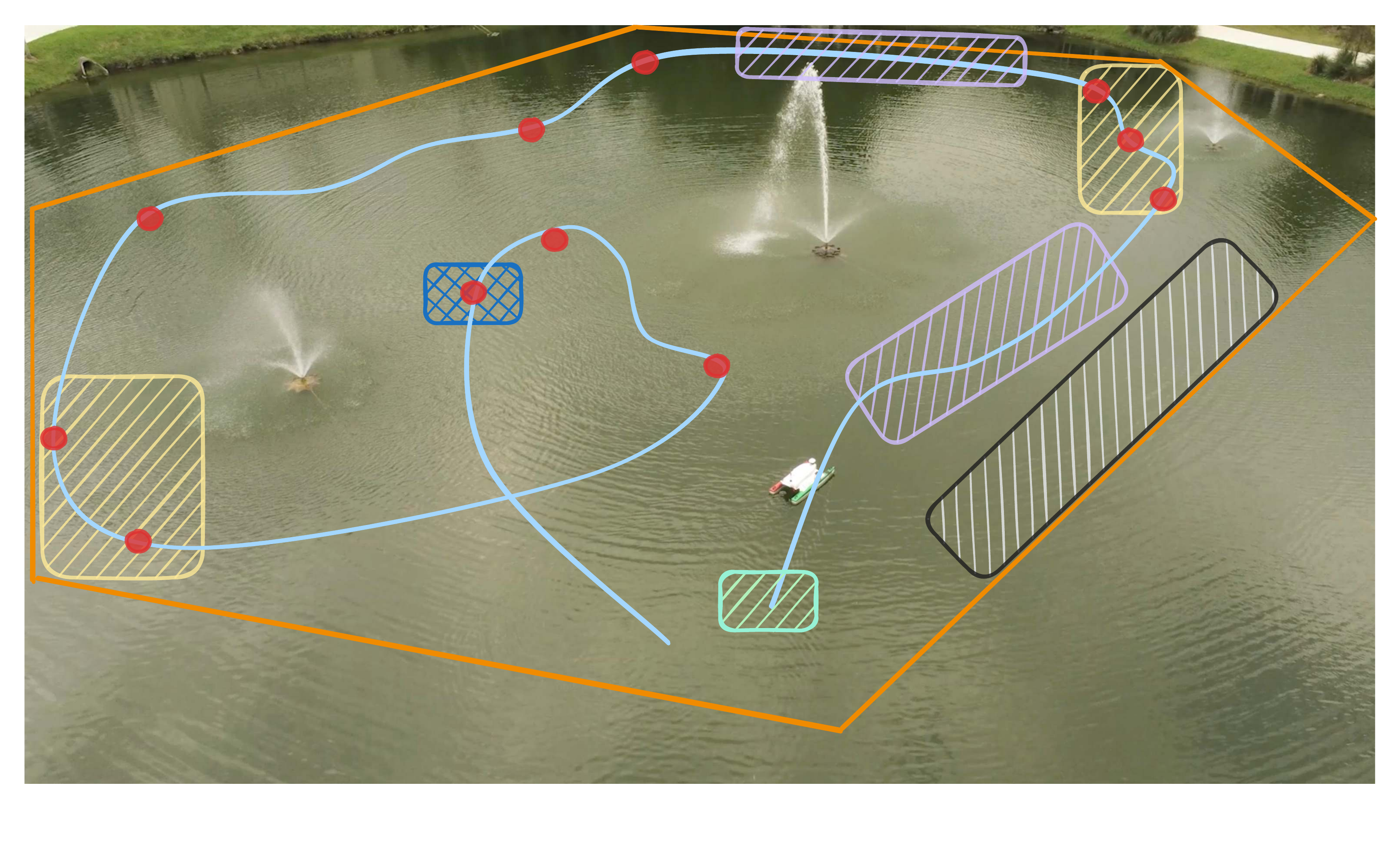}
        \caption*{\small GUIDEd SAC}
    \end{minipage}\\[2mm] 
    \begin{minipage}{\columnwidth}
        \centering
        \includegraphics[trim=45pt 130pt 20pt 30pt, clip, width=0.6\columnwidth]{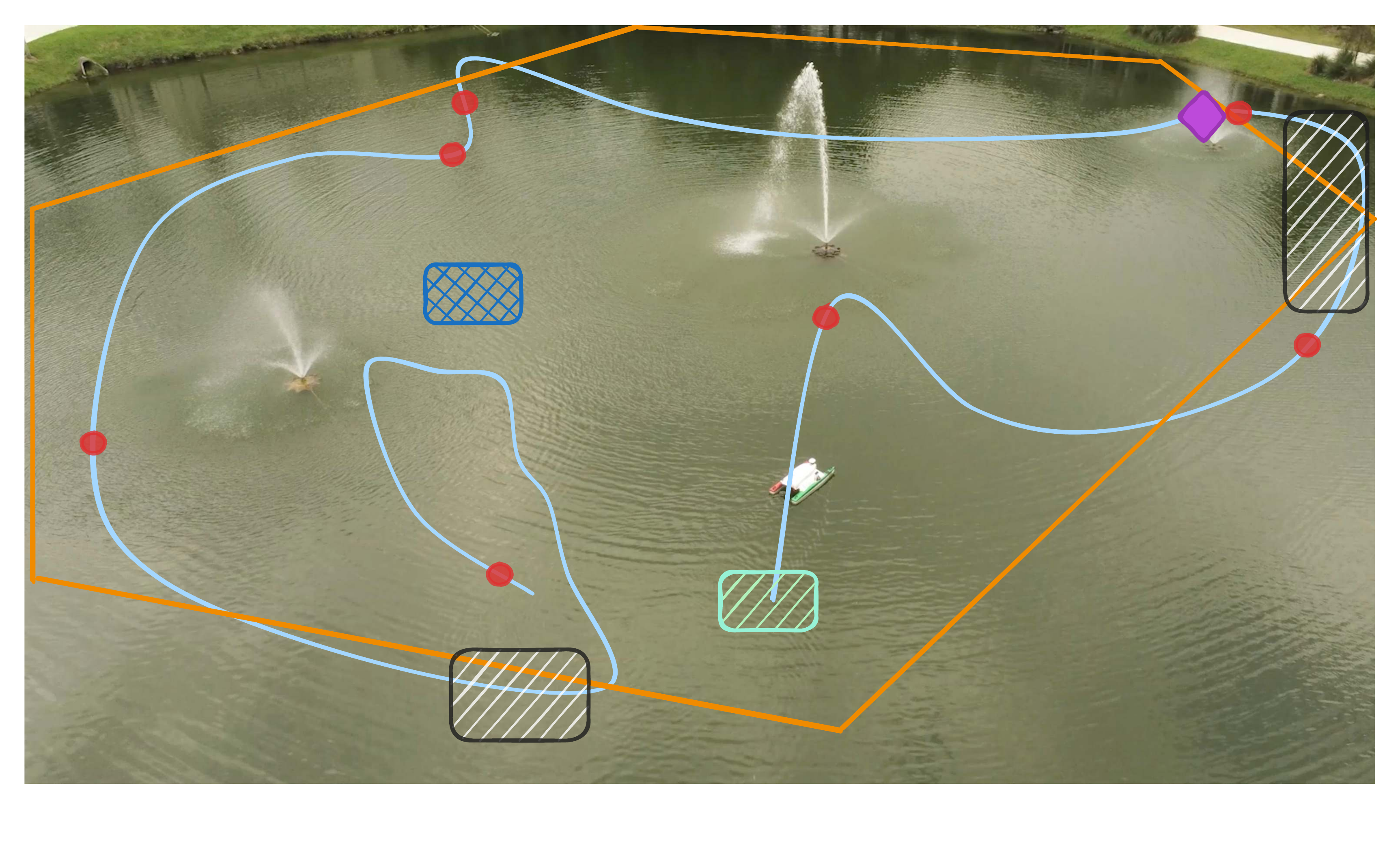}
        \caption*{\small SAC-P}
    \end{minipage}
    \caption{\small Trajectories for
\textcolor{guidelineBlue}{\texttt{Start and end at the dock. Go around the perimeter and visit (x,y).}} Green and blue rectangles denote the dock and the coordinates (x,y).}
    \label{fig:res2}
\end{figure}

\noindent \tbm{Comparison with Baselines:}
\label{sec:why-guided-wins}
Across settings in Table~\ref{tab:comparison}, the margin of improvement for GUIDE comes from two design choices that the baselines lack:
(i) \emph{spatially heterogeneous tolerance} via TSUM---the policy is trained to respect $u(s_t)\!\le\!U^\tau(s_t)$ where precision matters and to ignore uncertainty where it does not and 
(ii) \emph{conditioning rather than scalarizing}---we keep $U^\tau$ and $u$ as explicit inputs instead of folding them into a single reward term, which preserves gradients and avoids conflicting objectives (task progress vs.\ localization cost).
\emph{SAC} cannot reason about uncertainty tolerance at all (no $U^\tau$, no $u$), so it both overuses exact fixes in permissive areas and \emph{fails} to invoke them where precision is required (see Fig.~\ref{fig:results1}). 
\emph{SAC-P} applies a {uniform} penalty $-\zeta\,u(s)$ everywhere. This forces a single global trade-off that is simultaneously too conservative in open water and too lax at tight turns, yielding lower TCR.
\emph{B-SAC} estimates {epistemic} uncertainty to guide exploration, but our problem is dominated by {state-estimation} uncertainty; the signal it optimizes is misaligned with when precision is {valuable} for the task, so exploration is wasted where $U^\tau$ would not justify it.
\emph{CVaR-RL} optimizes worst-case returns, which induces \emph{global} risk-aversion. It forgoes beneficial, localized exact fixes and instead detours broadly.
\emph{RAA} minimizes overall collision probability without task semantics. It often chooses long, low-uncertainty paths that ignore the language-specified objective.
\emph{HEU} triggers exact localization by proximity to obstacles, but lacks semantic context and tolerance thereby failing to generalize in previously unseen settings.
The combination of a \emph{task-specific spatial tolerance} (TSUM) and \emph{explicit uncertainty input} gives GUIDE a mechanism the baselines lack: it turns ``where and when does precision matter?" into a learnable control problem which explains the consistent TCR and reward gains across all tasks.

\noindent \tbm{Behavior of GUIDEd Agents:}
Analyzing the specific task shown in Fig.~\ref{fig:res2}, GUIDEd agents strategically adjust their reliance on precise position estimation versus noisier estimates. In areas where the TSUMs indicate high precision is necessary--such as navigating near obstacles or close to the perimeters (Fig.~\ref{fig:res2}, highlighted in yellow)--the ASV opt for exact positioning despite the higher operational cost. Conversely, in less critical regions (Fig.~\ref{fig:res2}, highlighted in purple), they rely on less precise, cheap estimates. This adaptability allows GUIDEd agents to manage uncertainty more efficiently than baselines, resulting in better task completion rates.
Although not perfect, occasionally missing sections of the perimeter (Fig.~\ref{fig:res2}, highlighted in black), GUIDEd agents significantly outperform baselines with engineered rewards like SAC-P.
Baselines frequently fail to complete tasks safely, colliding with obstacles (Fig.~\ref{fig:res2}, pink diamond) or take inefficient paths (Fig.~\ref{fig:res2}, highlighted in black).

\section{Conclusion}
We introduce GUIDE, a framework for limited-localization environments that integrates task-specific uncertainty requirements into robotic navigation policies. Central to our approach is the concept of Task-Specific Uncertainty Maps (TSUMs), which represent acceptable levels of state estimation uncertainty across the environment based on the given task. By conditioning navigation policies on TSUMs, we enable robots to reason about the context-dependent importance of certainty.
We demonstrate how GUIDE can be incorporated into RL frameworks by augmenting the state representation with TSUMs. 
Specifically, we adapt SAC to operate in this augmented state space. Our experiments demonstrate GUIDE's effectiveness in real-world deployments across previously unseen environments, showing significant improvements in task completion rates compared to baselines that do not consider task-specific uncertainty.

\bibliographystyle{IEEEtran}
\bibliography{ref}

\end{document}